\documentclass[11pt,a4paper]{article}
\usepackage[utf8]{inputenc}
\usepackage[T1]{fontenc}
\usepackage{microtype}
\usepackage{lmodern}
\usepackage[english]{babel}
\usepackage{geometry}
\usepackage{amsmath,amssymb}
\usepackage{booktabs}
\usepackage{array}
\usepackage{hyperref}
\usepackage{enumitem}
\usepackage{xcolor}
\usepackage{graphicx}
\usepackage{float}
\usepackage{tikz}
\usetikzlibrary{positioning, arrows.meta, shapes.geometric, fit, backgrounds}
\usepackage{pgfplots}
\pgfplotsset{compat=1.18}
\usepackage{amsthm}

\newtheorem{proposition}{Proposition}
\usepackage{caption}
\usepackage{fancybox}
\usepackage{tcolorbox}
\usepackage{placeins}

\hypersetup{
  colorlinks=true,
  linkcolor={red!50!black},
  citecolor={blue!80!black},
  urlcolor={blue!80!black},
  pdftitle={Retrieval Is Not Enough: AI for Organizations Needs Epistemic Infrastructure},
  pdfauthor={Federico Bottino, Carlo Ferrero, Nicholas Dosio, Pierfrancesco Beneventano},
  pdfsubject={Epistemic memory and typed graph retrieval for organizational AI}
}

\newcolumntype{L}[1]{>{\raggedright\arraybackslash}p{#1}}
\newcolumntype{C}[1]{>{\centering\arraybackslash}p{#1}}

\title{\textbf{Retrieval Is Not Enough:\\AI for Organizations Needs Epistemic Infrastructure}}

\author{%
Federico Bottino$^{1}$, \quad Carlo Ferrero$^{1}$, \quad Nicholas Dosio$^{1}$\\
Pierfrancesco Beneventano$^{2}$\\[0.2cm]
\textit{$^{1}$Kakashi Ventures Accelerator (KVA)}\\
\textit{$^{2}$Massachusetts Institute of Technology}
}

\date{}

\begin{document}

\maketitle


\begin{abstract}
Retrieval-based organizational AI usually ranks passages by semantic relevance. In organizations, this is too weak: a passage can be relevant yet misleading because it records a superseded plan, a tentative hypothesis, an unresolved question, or one side of a later contradiction. We introduce OIDA, an epistemic memory architecture that stores these distinctions as persistent state rather than leaving them to prompt-time inference.

Formally, OIDA is a typed, attributed, signed, time-indexed graph. Nodes represent decisions, evidence, plans, hypotheses, constraints, and open information needs; edges encode support, contradiction, dependency, and supersession; retrieval combines semantic similarity with graph state. We also release three organizational epistemic-stress corpora for testing whether systems recover current decisions, contradictions, superseded claims, and knowledge gaps. The pilot evidence is preliminary: full-context baselines win in-window aggregate quality, while structured retrieval uses far fewer input tokens and its clearest positive signal is explicit knowledge-gap surfacing. The central claim is not that graph memory universally beats long context, but that organizational AI needs memory layers that represent epistemic state, not only text relevance.
\end{abstract}


\section{Introduction}
\label{sec:intro}

An AI agent is asked to summarize an organization's position on a strategic decision. It retrieves several relevant records: an early proposal, a later decision note, a recent market update that challenges the plan, and an unanswered regulatory question. All of them are topically relevant. Yet they do not play the same epistemic role. If the system treats them as interchangeable evidence, the resulting summary may be fluent and textually grounded while still misrepresenting the organization's actual state.

This kind of error is not primarily a failure of semantic retrieval. It is a failure of \emph{epistemic representation}. Organizational corpora do not contain only facts; they contain decisions, constraints, plans, hypotheses, evidence, observations, contradictions, and unresolved questions. For downstream agents, these distinctions are operational. A correct answer often depends not only on what a record says, but on what role that record plays in the organization's current state.

Existing work has substantially improved retrieval over enterprise and agent-memory corpora through better embeddings, reranking, graph-structured retrieval, and larger context windows. These advances help systems find relevant material. They do not by themselves make commitment, contradiction, supersession, or unresolved risk persistent and queryable properties of the memory layer. If unresolved questions and superseded claims are stored only as ordinary text, an agent must infer their status again at query time.

We introduce OIDA, an epistemic memory architecture for organizational AI. OIDA's core aim is to move epistemic state from transient prompt-time interpretation into persistent memory. Formally, we describe OIDA as a typed, attributed, signed, time-indexed knowledge graph. Nodes are typed memory objects: decisions, constraints, plans, hypotheses, observations, evidence items, evaluations, narratives, and open information needs. Edges are weighted relations such as \textsc{supports}, \textsc{contradicts}, \textsc{supersedes}, or \textsc{implements}. Retrieval can then combine semantic similarity with graph state: node type, provenance, timestamp, confidence, edge structure, and optional time-dependent priority weights.

Three design choices define the architecture. First, epistemic type matters: a decision should not be treated like an observation, and a hypothesis should not be treated like evidence. Second, contradictions should be represented as persistent signed relations, not rediscovered from raw text at query time. Third, unresolved questions should be first-class memory objects, because an organization's unknowns can be as operationally important as its settled facts.

The question is not whether graph-structured memory always beats dense retrieval or long-context prompting. It does not. If the full relevant corpus fits in context at acceptable cost, a long-context baseline is often the strongest reference point. The relevant question is diagnostic: for which organizational datasets does semantic retrieval fail because the answer depends on current commitment, contradiction, supersession, or unresolved risk rather than topical relevance alone?

This paper addresses that question in two ways. First, it specifies an epistemic memory substrate that makes organizational state explicit enough to retrieve. Second, it releases organizational epistemic-stress corpora designed to test whether systems recover that state. The experiments are preliminary and intentionally diagnostic. They show that full-context baselines win in the in-window regime, while selective graph retrieval uses far fewer input tokens and can surface knowledge gaps, but is vulnerable to incomplete extraction or retrieval. This turns the method question into a dataset question: graph memory is worth considering when the corpus and query distribution make epistemic state hard to recover from relevance alone.

\paragraph{Contributions.}
\begin{enumerate}[nosep,leftmargin=*]
\item \textbf{Epistemic memory architecture.}
We introduce OIDA, an architecture for representing organizational knowledge as persistent epistemic state: decisions, evidence, hypotheses, plans, constraints, contradictions, supersessions, and open information needs.

\item \textbf{Typed graph formalization.}
We formalize the architecture as a typed, attributed, signed, time-indexed graph whose nodes are memory objects and whose edges encode support, dependency, contradiction, and supersession. This lets retrieval condition semantic relevance on epistemic state.

\item \textbf{Organizational epistemic-stress corpora.}
We release three synthetic but structurally realistic corpora spanning consulting operations, IoT product development, and venture-capital deliberation, designed to test epistemic retrieval failures rather than topical relevance alone.

\item \textbf{Pilot evaluation and falsification path.}
We report pilot evidence for knowledge-gap surfacing and a clear in-window result: full-context baselines dominate aggregate quality when the corpus fits in context. We state the remaining decisive tests: static typed-graph baselines, contradiction-edge precision, and longitudinal node-weight telemetry.
\end{enumerate}

\paragraph{Evidence and scope.}
The representation and update rule are architectural specifications. The corpora are released artifacts. The pilot observations support a narrow behavioral claim: explicit graph state can change what a system surfaces, especially knowledge gaps. We do not claim longitudinal dynamics, dynamic-vs-static advantage, or out-of-window superiority. Those are next tests, not results of this paper.

The remainder of the paper proceeds as follows. Section~\ref{sec:related} positions the work. Section~\ref{sec:graph} gives the graph formulation. Section~\ref{sec:corpora} describes the released corpora. Section~\ref{sec:eval} reports the pilot evaluation. Section~\ref{sec:when-to-use} gives dataset-suitability criteria. Section~\ref{sec:limitations} states what is not yet established.


\section{Related Work}
\label{sec:related}

\paragraph{Retrieval and long context.}
Retrieval-augmented generation grounds model outputs in external text~\cite{lewis2020,gao2024ragsurvey}. Dense retrieval, reranking, graph-augmented retrieval~\cite{edge2024graphrag,guo2024lightrag}, and long-context prompting all improve access to relevant material. The distinction made here is narrower: topical access is not the same as recovering epistemic state. A system may retrieve a relevant note and still miss that it is a draft, has been superseded, or contradicts a later document.

\paragraph{Knowledge graphs and temporal state.}
Knowledge graphs represent entities and typed relations~\cite{bernerslee2001}; temporal knowledge graphs add validity intervals and time-aware reasoning~\cite{cai2024tkg}. These tools are natural analogues for organizational memory. The missing distinction for our setting is role, not only time: a memory item may be a decision, plan, evidence item, hypothesis, observation, or open information need, and those roles should affect retrieval.

\paragraph{Agent memory.}
Agent-memory systems such as MemGPT~\cite{packer2023}, Mem0~\cite{chhikara2025mem0}, Zep/Graphiti~\cite{rasmussen2025zep}, MemOS~\cite{memos2025}, and A-MEM~\cite{xu2025amem} show that agents benefit from external memory, hierarchy, and consolidation. The present paper studies a complementary question: when should memory items carry explicit epistemic type and signed graph state so that retrieval can distinguish current commitments from tentative or contradicted material?

\paragraph{Contradiction, provenance, and belief update.}
Provenance standards such as PROV-O~\cite{provo2013} make source lineage explicit. Signed graphs provide a standard language for positive and negative relations~\cite{signedGNNsurvey2024}. Work on knowledge conflicts and inconsistency studies how systems behave when context, memory, or sources disagree~\cite{xu2024knowledgeconflicts,kgInconsistencySurvey2025}. We use these ideas operationally: contradiction is represented as graph structure that can affect retrieval priority, but the present experiments do not yet validate contradiction-edge precision or longitudinal belief update.

\paragraph{Organizational memory.}
Organizational memory theory emphasizes that knowledge is distributed across people, processes, documents, culture, and routines~\cite{walsh1991,stein1995,nonaka1994}. The graph formulation in this paper is a computational substrate for one part of that problem: representing the status of organizational records so downstream agents can tell what is known, what is binding, what is stale, and what remains unresolved.


\section{OIDA: Epistemic Memory as Typed Graph State}
\label{sec:graph}

OIDA is an epistemic memory architecture. Its implementation has product-specific details, but the core scientific object is simple: organizational memory at time $t$ is an attributed directed graph
\[
G_t = (V_t, E_t, X_t, R_t, K_t).
\]
Each node $v \in V_t$ is a typed memory object: a decision, constraint, evidence item, plan, hypothesis, observation, evaluation, narrative, or open information need. Node attributes $X_t(v)$ include source content, provenance, timestamp, epistemic type, and optional confidence metadata. Each edge $e=(u,v) \in E_t$ has a relation type and weight, with positive relations such as \textsc{supports} and \textsc{implements} and negative or inhibitory relations such as \textsc{contradicts} and \textsc{blocks}. The scalar $K_t(v)$ is a node-priority weight used for retrieval.

The graph language is not a retreat from the architectural claim. It is the way we make the claim precise: organizational AI needs memory state that records what a piece of information is, how it relates to other information, and whether it should still guide action. In the implementation, typed nodes are called Knowledge Objects; in the paper, we mostly use the graph terms because they are more general.

\begin{figure}[H]
\centering
\begin{tikzpicture}[
  node distance=5mm and 7mm,
  every node/.style={font=\small},
  box/.style={rectangle, draw, rounded corners=3pt, text width=2.15cm, minimum height=8mm, fill=blue!5, align=center},
  graphbox/.style={rectangle, draw, rounded corners=3pt, text width=2.45cm, minimum height=8mm, fill=blue!5, align=center},
  arr/.style={-{Stealth[length=5pt]}, semithick}
]
\node[box] (docs) {Documents};
\node[box, right=of docs] (nodes) {Typed memory nodes};
\node[graphbox, right=of nodes] (graph) {Signed graph state $G_t$};
\node[box, right=of graph] (retr) {Retrieval over text + graph};
\node[box, right=of retr] (ctx) {Answer context};
\draw[arr] (docs) -- node[above,font=\scriptsize]{ingest} (nodes);
\draw[arr] (nodes) -- node[above,font=\scriptsize]{edges} (graph);
\draw[arr] (graph) -- node[above,font=\scriptsize]{rank} (retr);
\draw[arr] (retr) -- node[above,font=\scriptsize]{select} (ctx);
\draw[arr,dashed] (ctx.south) -- ++(0,-7mm) -| node[below,font=\scriptsize,pos=.25]{usage / revision signals} (graph.south);
\end{tikzpicture}
\caption{OIDA as epistemic memory represented in graph form. Nodes carry epistemic type, provenance, and temporal metadata; edges carry relation type and weight; retrieval combines semantic similarity with graph state.}
\label{fig:graph-memory}
\end{figure}

\subsection{Memory Objects as Typed Nodes}
\label{sec:nodes}

A memory object is a compact representation of a source-backed organizational unit. It is not intended to replace the raw document. It adds typed state that raw text alone does not provide reliably at retrieval time. In the implementation these units are Knowledge Objects; in the formal model they are typed nodes in the memory graph.

\[
v_i = (\mathrm{id}_i,\; \tau_i,\; \mathrm{type}_i,\; \mathrm{content}_i,\; \mathrm{prov}_i,\; \mathrm{scores}_i,\; \mathrm{meta}_i),
\]
where $\tau_i$ is temporal metadata, $\mathrm{type}_i$ is the epistemic type, $\mathrm{prov}_i$ is provenance, and $\mathrm{scores}_i$ may include confidence, freshness, urgency, contradiction status, and retrieval priority.

\begin{table}[H]
\centering
\footnotesize
\caption{Epistemic node types. Initial priorities and temporal behaviors are design priors, not validated optima.}
\label{tab:types}
\begin{tabular}{@{}L{2.5cm}C{1.4cm}L{3.1cm}L{4.9cm}@{}}
\toprule
\textbf{Type} & \textbf{Initial priority} & \textbf{Temporal behavior} & \textbf{Role} \\
\midrule
DECISION    & 1.00 & Stable until superseded & Binding choice or commitment \\
CONSTRAINT  & 0.90 & Stable until changed & Hard structural boundary \\
EVIDENCE    & 0.80 & Slow decay & Verifiable supporting or refuting material \\
NARRATIVE   & 0.70 & Stable contextual anchor & Persistent interpretive context \\
PLAN        & 0.65 & Time-bounded decay & Structured intention with horizon \\
EVALUATION  & 0.55 & Moderate decay & Informed qualitative assessment \\
OBSERVATION & 0.40 & Unreinforced decay & Weak uninterpreted signal \\
HYPOTHESIS  & 0.30 & Decay if untested & Unverified testable claim \\
QUESTION    & 0.30 & Urgency grows while unresolved & Open information need \\
\bottomrule
\end{tabular}
\end{table}

The type list is not a claim about human cognition. It is an engineering schema for retrieval. A correct answer to ``what is our current position?'' should treat a decision differently from a draft plan, and an open information need differently from a settled fact.

\subsection{Signed and Weighted Edges}
\label{sec:edges-main}

Edges encode relations between memory nodes. Positive edges such as \textsc{supports}, \textsc{implements}, and \textsc{based\_on} provide reinforcing structure. Negative or inhibitory edges such as \textsc{contradicts} and \textsc{blocks} mark tension that should not be hidden by topical similarity. The important design choice is non-destructive contradiction handling: a contradicted node is down-weighted or paired with its contradiction, not deleted. This matters because organizational contradictions are often unresolved coexistences rather than immediate logical defeats.

\subsection{Node-Priority Weights}
\label{sec:priority}

A node-priority weight $K_t(v)$ can be computed from type, time, use, evidence, graph influence, and contradiction. In the present paper this is a design mechanism, not a measured longitudinal result. The following update rule specifies the intended dynamics:
\begin{equation}
K_{t+1}(v)
=
\mathrm{clamp}_{[0,1]}\!\Big(
(1-\eta)K_t(v)
+
\eta\,[s(v)+h_t(v)+u_t(v)+e_t(v)+g_t(v)]
-
\lambda_{\mathrm{type}(v)}\Delta t\,K_t(v)
-
c_t(v)
\Big).
\label{eq:priority-update}
\end{equation}
Here $s(v)$ is the type prior, $h_t(v)$ is an urgency term for open information-need nodes and zero otherwise, $u_t(v)$ is usage activation, $e_t(v)$ is evidence/support activation, $g_t(v)$ is graph influence, and $c_t(v)$ is a contradiction penalty.

Equation~\ref{eq:priority-update} is a model specification, not an empirical result in this paper. We do not report longitudinal telemetry showing node priorities evolving over repeated update cycles.

\begin{proposition}[Open-question urgency threshold]
Let an open information-need node have base priority $s_Q=0.30$ and urgency term $h(t)\geq 0$, and let an observation node have base priority $s_O=0.40$. Under stationary inputs $(u=e=g=c=0)$, $\eta=0.1$, $\Delta t=1$, $\lambda_Q=-0.010$, and $\lambda_O=0.015$, the fixed points are
\[
K_Q^*(t)=\frac{\eta(s_Q+h(t))}{\eta+\lambda_Q\Delta t},
\qquad
K_O^*=\frac{\eta s_O}{\eta+\lambda_O\Delta t}.
\]
Then $K_Q^*(t)>K_O^*$ whenever $h(t)>0.013$. Thus unresolved questions outrank unreinforced observations only once the urgency term crosses a small positive threshold; the rising-priority behavior is a property of the urgency mechanism, not of the label alone.
\end{proposition}

\subsection{Retrieval over Text and Graph State}
\label{sec:retrieval}

Retrieval ranks nodes by combining semantic similarity with graph state:
\begin{equation}
\mathrm{score}(q,v,t) = \mathrm{sim}(q,v) \cdot \phi\!\left(K_t(v),\mathrm{type}(v),E_t(v),q\right),
\label{eq:graph-retrieval}
\end{equation}
where $\mathrm{sim}(q,v)$ is semantic similarity between query and node content, and $\phi$ is a bounded modulation term based on priority, node type, local edges, and query context. The exact choice of $\phi$ is implementation-specific. The point is architectural: relevance should be conditioned on what kind of memory item is being retrieved.


\section{Organizational Epistemic-Stress Corpora}
\label{sec:corpora}

Standard retrieval benchmarks usually test topical relevance. Organizational AI also needs corpora that test epistemic-state recovery: current decisions, unresolved questions, superseded plans, contradictions, and uncertain evidence. We therefore release three synthetic but structurally realistic organizational corpora.

\begin{table}[H]
\centering
\small
\caption{Released organizational epistemic-stress corpora. The public repository describes the corpora as synthetic but structurally realistic organizational knowledge bases.}
\label{tab:corpora}
\begin{tabular}{@{}L{2.4cm}L{3.0cm}C{2.1cm}L{6.0cm}@{}}
\toprule
\textbf{Corpus} & \textbf{Domain} & \textbf{Files / size} & \textbf{Intended stress tests} \\
\midrule
A / ClearPath & Consulting / operations & 46 / 318 KB & Process redesign, bottleneck evidence, unresolved audit workflow, and conflicting organizational observations \\
B / FireGlass & IoT / product development & 47 / 566 KB & Architecture decisions, latency constraints, technical debt, negative knowledge, and integration risk \\
C / Vertex Minds & Venture capital & 77 / 376 KB & Investment decisions, due-diligence gaps, prior-deal lessons, and market-size contradiction \\
\bottomrule
\end{tabular}
\end{table}

Each corpus follows an eight-category source schema: project scope, domain material, internal communications, external communications, meetings, market context, formal documents, and agenda/calendar artifacts. The corpora are designed to contain diverse epistemic situations: binding decisions, open questions, contested hypotheses, contradictory evidence, and evolving plans. The corpus files and evaluation artifacts are available in the \href{https://github.com/kakashi-ventures/oida-evaluation-corpora}{OIDA evaluation corpora repository}~\cite{oidacorpora2026}.

The dataset contribution is separable from the graph method. The corpora are meant to be reusable stress tests for epistemic retrieval, not demonstrations of one implementation. They can test ordinary RAG, long-context prompting, graph-augmented retrieval, typed graph retrieval, or static labeling baselines. The intended test is not only ``did the system retrieve a relevant chunk?'' but ``did it recover the current epistemic state?'' Because the organizations, people, and events are fictional, the corpora support controlled release and adversarial reuse, but they should not be confused with natural deployment logs.


\section{Pilot Evaluation: In-Window Quality vs. Context Cost}
\label{sec:eval}

The evaluation is a pilot, not a benchmark-scale validation. It asks a regime question: what happens when selective structured retrieval is compared against a full-context baseline while the whole corpus fits in context? This is a strong regime for full context because the baseline sees all source material. The point is to measure the quality--cost tradeoff and expose failure modes, not to claim that graph retrieval wins in-window.

The evaluation can answer three limited questions: how large the in-window full-context advantage is, how much less input context selective retrieval uses, and which epistemic failure modes appear when retrieval is incomplete. It cannot answer whether time-dependent node weights improve retrieval; that requires a static typed-graph baseline and longitudinal telemetry. The reported corpus evaluations use the fixed generator--judge protocol recorded in the corpus reports: three queries per corpus, ten response runs per condition, structured retrieval versus full context. The numbers should be treated as within-bench diagnostics rather than externally calibrated measurements.

\begin{table}[H]
\centering
\small
\renewcommand{\arraystretch}{1.18}
\caption{Evaluation scope in this draft. The release contains three corpora; the reported analysis uses Corpus A and Corpus C.}
\label{tab:evaluation-scope}
\begin{tabular}{@{}L{2.3cm}L{2.7cm}L{5.2cm}L{3.6cm}@{}}
\toprule
\textbf{Corpus} & \textbf{Status here} & \textbf{Use in this paper} & \textbf{Not established} \\
\midrule
A / ClearPath & Reported pilot & Aggregate in-window comparison; knowledge-gap surfacing & No longitudinal dynamics \\
\addlinespace[2pt]
B / FireGlass & Released dataset & Product-development stress corpus for future tests & Not analyzed here \\
\addlinespace[2pt]
C / Vertex Minds & Reported diagnostic & Quality--cost frontier; contradiction-sensitive failure mode & No graph-retrieval win \\
\bottomrule
\end{tabular}
\end{table}

\subsection{Evaluation Rubric}

We use the Epistemic Quality Score (EQS) as a diagnostic rubric with five dimensions:
\[
\mathrm{EQS} = 0.20\,\mathrm{ECA} + 0.25\,\mathrm{CP} + 0.20\,\mathrm{CR} + 0.20\,\mathrm{EC} + 0.15\,\mathrm{DE}.
\]
The dimensions are type fidelity (ECA), groundedness (CP), context coverage (CR), contradiction handling (EC), and decision utility (DE). EQS is not a validated community metric. The numbers below use the LLM-judge protocol from the corpus reports; since LLM judges can have systematic biases~\cite{zheng2023judge}, we read them as within-bench diagnostics rather than external truth.

\subsection{Corpus A / ClearPath: In-Window Aggregate}

Corpus A is a consulting / operations corpus with three queries. The full-context condition has higher aggregate EQS, and the largest gap is context coverage. This is the expected in-window pattern: the full-context baseline sees the whole corpus, while structured retrieval receives a small selected context.

\begin{table}[H]
\centering
\small
\caption{Corpus A / ClearPath aggregate EQS over three queries. The largest gap is contextual recall, so the result is dominated by breadth rather than a clean test of graph structure.}
\label{tab:clearpath-eqs}
\begin{tabular}{@{}lcccr@{}}
\toprule
\textbf{Sub-score} & \textbf{Weight} & \textbf{Structured} & \textbf{Full context} & $\boldsymbol{\Delta}$ \\
\midrule
ECA & 0.20 & 0.667 & 0.807 & +0.140 \\
CP  & 0.25 & 0.550 & 0.862 & +0.312 \\
CR  & 0.20 & 0.418 & 0.860 & +0.442 \\
EC  & 0.20 & 0.578 & 0.788 & +0.210 \\
DE  & 0.15 & 0.582 & 0.843 & +0.262 \\
\midrule
\textbf{EQS} & 1.00 & \textbf{0.5574} & \textbf{0.8329} & \textbf{+0.2755} \\
\bottomrule
\end{tabular}
\end{table}

\begin{table}[H]
\centering
\small
\caption{Corpus A quality--cost summary. Structured retrieval is lower quality in this regime but uses far fewer input tokens.}
\label{tab:clearpath-cost}
\begin{tabular}{@{}lccr@{}}
\toprule
\textbf{Metric} & \textbf{Structured} & \textbf{Full context} & \textbf{Ratio / gap} \\
\midrule
Mean input tokens & 2{,}623 & 110{,}539 & 42.1$\times$ \\
Aggregate EQS & 0.5574 & 0.8329 & +0.2755 \\
Contextual recall & 0.418 & 0.860 & +0.442 \\
\bottomrule
\end{tabular}
\end{table}

A separate single-query ClearPath pilot measured explicit knowledge-gap surfacing. Structured retrieval included a dedicated Knowledge Gap section in all 10 runs; the full-context baseline surfaced the same limitation in 5 of 10 runs. This is the clearest positive behavioral signal, but it should be read narrowly. It suggests that representing open information needs can change answer behavior. It does not show that time-dependent priority weights caused the behavior, and it does not validate longitudinal dynamics.

\begin{table}[H]
\centering
\small
\caption{ClearPath knowledge-gap surfacing. This is the pilot's clearest positive signal, but it remains a small-$n$ behavioral observation.}
\label{tab:knowledge-gap}
\begin{tabular}{@{}L{3.4cm}C{2.0cm}L{4.9cm}@{}}
\toprule
\textbf{Condition} & \textbf{Gap section} & \textbf{Interpretation} \\
\midrule
Structured graph retrieval & 10 / 10 & Open questions are foregrounded by the representation \\
Full-context baseline & 5 / 10 & Gaps appear when salient in the raw context \\
\bottomrule
\end{tabular}
\end{table}

\subsection{Corpus C / Vertex Minds: Contradiction Failure Mode}

Vertex Minds is a diagnostic check, not a second success claim. Corpus C contains 77 documents and tests investment-committee decision-making. The protocol uses 10 runs for each of two conditions across three queries. The full-context baseline again has higher aggregate EQS, especially context coverage. We include it because it makes the quality--cost tradeoff and a contradiction-sensitive failure mode visible in one corpus.

\begin{table}[H]
\centering
\small
\caption{Corpus C / Vertex Minds aggregate EQS. The largest gap is contextual recall, consistent with the full-context baseline's breadth advantage.}
\label{tab:corpus-c-eqs}
\begin{tabular}{@{}lcccr@{}}
\toprule
\textbf{Sub-score} & \textbf{Weight} & \textbf{Structured} & \textbf{Full context} & $\boldsymbol{\Delta}$ \\
\midrule
ECA & 0.20 & 0.617 & 0.817 & +0.200 \\
CP  & 0.25 & 0.550 & 0.865 & +0.315 \\
CR  & 0.20 & 0.455 & 0.863 & +0.408 \\
EC  & 0.20 & 0.548 & 0.822 & +0.273 \\
DE  & 0.15 & 0.520 & 0.848 & +0.328 \\
\midrule
\textbf{EQS} & 1.00 & \textbf{0.5395} & \textbf{0.8438} & \textbf{+0.3043} \\
\bottomrule
\end{tabular}
\end{table}

\begin{table}[H]
\centering
\scriptsize
\caption{In-window quality--cost summary for the two corpus reports analyzed here. Full context is higher quality; structured retrieval uses much less input context.}
\label{tab:quality-cost-summary}
\begin{tabular}{@{}lcccr@{}}
\toprule
\textbf{Corpus} & \textbf{Struct. tok.} & \textbf{Full tok.} & \textbf{Struct. EQS} & \textbf{Full EQS} \\
\midrule
A / ClearPath & 2{,}623 & 110{,}539 & 0.5574 & 0.8329 \\
C / Vertex Minds & 1{,}944 & 109{,}303 & 0.5395 & 0.8438 \\
\bottomrule
\end{tabular}
\end{table}

\begin{figure}[H]
\centering
\begin{tikzpicture}
\begin{axis}[
  width=0.88\linewidth,
  height=5.4cm,
  xlabel={Mean input tokens (log scale)},
  ylabel={Aggregate EQS},
  xmode=log,
  xmin=1200, xmax=160000,
  ymin=0.48, ymax=0.88,
  xtick={2000,4000,10000,100000},
  xticklabels={2k,4k,10k,100k},
  ytick={0.5,0.6,0.7,0.8},
  grid=major,
  legend style={font=\small, at={(0.02,0.98)}, anchor=north west}
]
\addplot+[only marks, mark=*] coordinates {(2623,0.5574) (1944,0.5395)};
\addlegendentry{Structured retrieval}
\addplot+[only marks, mark=square*] coordinates {(110539,0.8329) (109303,0.8438)};
\addlegendentry{Full context}
\end{axis}
\end{tikzpicture}
\caption{In-window quality--cost comparison for Corpus A and Corpus C. Full-context access is higher quality in this regime, but uses much more input context. The figure diagnoses the tradeoff rather than claiming a win for structured retrieval.}
\label{fig:quality-cost}
\end{figure}

\begin{figure}[H]
\centering
\begin{tikzpicture}
\begin{axis}[
  ybar,
  width=0.85\linewidth,
  height=5.0cm,
  ylabel={Full-context minus structured score},
  symbolic x coords={ECA,CP,CR,EC,DE},
  xtick=data,
  ymin=0, ymax=0.45,
  nodes near coords,
  nodes near coords align={vertical},
  grid=major,
  bar width=13pt
]
\addplot coordinates {(ECA,0.200) (CP,0.315) (CR,0.408) (EC,0.273) (DE,0.328)};
\end{axis}
\end{tikzpicture}
\caption{Corpus C sub-score gap decomposition. The largest gap is contextual recall, so the aggregate result is dominated by context breadth.}
\label{fig:subscore-gap}
\end{figure}

The complex Corpus C query exposes the central failure mode. The query asks for NovaTech AI's market size and reliability. Structured retrieval may present a single TAM figure and miss the pitch-deck vs. dossier gap. Full context recovers both figures and treats them as unverified hypotheses. This is not a reason to discard graph memory; it is a precise statement of what the graph layer must get right: both sides of a tension must be extracted and retrieved.

\begin{table}[H]
\centering
\small
\caption{Corpus C contradiction-sensitive query. The example shows why extraction and retrieval completeness are load-bearing for typed graph memory.}
\label{tab:tam-failure}
\begin{tabular}{@{}L{3.0cm}L{4.1cm}L{4.1cm}L{3.0cm}@{}}
\toprule
\textbf{Query} & \textbf{Structured retrieval behavior} & \textbf{Full-context behavior} & \textbf{Diagnostic lesson} \\
\midrule
NovaTech market size and reliability & May present one TAM figure and miss the pitch-deck vs. dossier contradiction & Recovers both figures and treats them as unverified hypotheses & Graph retrieval helps only if both sides of a tension are extracted and retrieved \\
\bottomrule
\end{tabular}
\end{table}

\paragraph{Interpretation.}
The pilot should not be read as evidence that structured graph retrieval outperforms full-context retrieval in the in-window regime. It does not. The full-context condition receives the entire corpus and achieves higher aggregate EQS. The useful result is diagnostic: the aggregate gap is driven primarily by context breadth, while the structured condition uses far fewer input tokens and, in one ClearPath pilot, surfaces knowledge gaps more consistently. The practical question is whether graph state improves the quality--cost frontier once full context is unavailable, too expensive, or too noisy. A positive next result would not be ``graph retrieval wins everywhere''; it would be narrower: typed graph retrieval improves current-state, contradiction, or open-question queries at equal token budget, and dynamic weights improve over a static typed graph. The next tests are a static typed-graph baseline, an out-of-window setting where selection is necessary, and extraction audits for contradiction and open-question nodes.

\FloatBarrier


\section{When Is Epistemic Graph Memory Worth Using?}
\label{sec:when-to-use}

Epistemic graph memory is not a universal replacement for dense retrieval or long-context prompting. It is useful when the answer depends on state that topical similarity alone does not reliably recover: what is decided, what is tentative, what has been contradicted, what has been superseded, and what remains unknown.

A simple adoption rule follows. Use full context if the relevant corpus fits in context at acceptable cost. Use ordinary RAG if queries are mostly single-hop factual lookups. Use an epistemic graph layer when the corpus contains evolving decisions, contradictory claims, open information needs, heterogeneous authority, or multi-document decision chains under a real context budget. The graph layer adds extraction and maintenance cost. It is worth paying that cost only when it buys state recovery that ordinary retrieval does not provide.

Before building a graph layer, an organization can run a small corpus diagnosis: sample documents, label realistic queries, and estimate whether the query distribution actually requires epistemic state. The output should be a decision procedure, not an intuition: if full context is affordable, use it; if ordinary RAG answers the query set, avoid graph overhead; if current-state, contradiction, or open-question queries fail, test graph retrieval. Table~\ref{tab:diagnostics-measures} lists the quantities that matter most.

\begin{table}[H]
\centering
\small
\caption{Pre-deployment diagnostics. These quantities estimate whether epistemic graph memory is likely to help on a deployment dataset.}
\label{tab:diagnostics-measures}
\begin{tabular}{@{}L{3.1cm}L{5.5cm}L{4.9cm}@{}}
\toprule
\textbf{Quantity} & \textbf{How to estimate it} & \textbf{Why it matters} \\
\midrule
Window pressure & Tokens in the task-relevant corpus divided by usable context budget & If pressure is low, full context may be best \\
Epistemic heterogeneity & Fraction of units that are decisions, plans, hypotheses, evidence, or open questions & Flat chunks are less harmful when all records play the same role \\
Conflict / supersession rate & Fraction of claim-like units that contradict or replace earlier units & Signed edges matter only when conflicts or changes are common \\
Open-question rate & Number of unresolved information needs per document or per query & Open-question nodes matter only if uncertainty must be surfaced \\
Multi-document dependency & Number of source documents needed for a correct answer & Graph state helps most when answers require chains or tensions \\
Extraction quality & Precision and recall of node types and edge types on a small audit set & Bad graph structure can be worse than no graph structure \\
\bottomrule
\end{tabular}
\end{table}

\begin{table}[H]
\centering
\small
\caption{Minimal adoption experiment. Each condition isolates a different source of value.}
\label{tab:adoption-experiment}
\begin{tabular}{@{}L{3.2cm}L{5.1cm}L{5.1cm}@{}}
\toprule
\textbf{Condition} & \textbf{What it tests} & \textbf{Decision rule} \\
\midrule
Full context & Best in-window answer quality when the corpus fits & Use if cost and latency are acceptable \\
Dense / hybrid RAG & Standard low-cost retrieval without typed graph state & Use if queries are mostly factual and single-hop \\
Static typed graph & Value of typed nodes and signed edges without time dynamics & Use if state-sensitive queries improve at equal budget \\
Dynamic typed graph & Added value of time-dependent node weights & Use only if it beats the static typed graph \\
\bottomrule
\end{tabular}
\end{table}

A practical gate follows from these diagnostics. If window pressure is low and full-context cost is acceptable, the graph layer is usually unnecessary. If extraction quality is poor, graph memory should not be deployed even when the corpus has contradictions or open questions. If a static typed graph performs as well as a dynamic one, the time-dependent update rule is not yet justified.

For a new organizational dataset, the decisive experiment should compare four conditions: full context where feasible, dense/hybrid RAG, static typed graph retrieval, and dynamic time-weighted graph retrieval. The graph layer is justified only if it improves contradiction, current-state, or open-question queries at equal or lower token budget. The dynamic layer is justified only if it improves over the static typed graph baseline. This turns adoption into a measurable decision, not a system preference.

\FloatBarrier


\section{Limitations}
\label{sec:limitations}

\paragraph{No longitudinal validation.}
The paper specifies time-dependent node-priority dynamics, but does not measure node trajectories over repeated update cycles. Type-specific decay, urgency accumulation, usage activation, and contradiction penalties are formal mechanisms here; they require longitudinal telemetry before being treated as empirical results.

\paragraph{No dynamic-vs-static isolation.}
The current experiments do not isolate whether observed behavior comes from typed graph representation, prompt formatting, retrieval selection, or time-dependent node weights. A static typed-graph baseline is the decisive comparison. Dynamic weighting should therefore be treated as an open hypothesis rather than an established result.

\paragraph{In-window regime.}
The full-context baseline sees the entire corpus and therefore wins aggregate answer quality in the reported setting. The experiments diagnose the quality-cost frontier; they do not show that graph retrieval beats long context when full context is available.

\paragraph{LLM-as-judge evaluation.}
The EQS scores are produced by an LLM judge. They are useful for consistent within-bench comparisons, but they are not a substitute for human expert assessment of organizational decision quality~\cite{zheng2023judge}.

\paragraph{Extraction quality is load-bearing.}
Typed graph retrieval only helps if relevant nodes and edges are extracted with adequate precision and recall. Missing one side of a contradiction can produce false consistency. A final system evaluation should therefore report node-type accuracy and edge-type precision, not only answer quality.

\paragraph{Synthetic corpora.}
The released corpora are synthetic but structurally realistic. They are useful for controlled epistemic stress tests, but natural organizational corpora remain necessary for deployment claims.


\section{Conclusion}
\label{sec:conclusion}

This paper studies a representational bottleneck in organizational AI. A system may retrieve semantically relevant organizational material and still produce an epistemically distorted answer if its memory layer does not distinguish decisions from hypotheses, resolved findings from open questions, or current claims from contradicted ones.

OIDA is a concrete attempt to make these distinctions part of persistent state rather than leaving them entirely to prompt-time interpretation. Formally, we represent that state as a typed, attributed, signed graph: memory objects carry epistemic type, provenance, temporal metadata, and priority weights; edges encode support, dependency, contradiction, and supersession; retrieval combines semantic similarity with graph state. We also release three organizational epistemic-stress corpora designed to expose failure modes that ordinary retrieval benchmarks often miss.

The experiments should be read with the right scope. In the in-window regime, full-context baselines win aggregate answer quality. Structured graph retrieval uses far fewer input tokens and can surface knowledge gaps, but it can also miss critical context when extraction or retrieval is incomplete. This does not erase the architectural claim; it sharpens it. The graph layer is valuable only when the corpus contains epistemic state that must be recovered under a real context budget: evolving decisions, contradictions, open questions, heterogeneous authority, and multi-document decision chains.

The broader conclusion is that organizational AI should be evaluated not only by how well it retrieves text, but also by whether the underlying memory layer represents commitment, contradiction, supersession, and unresolved ignorance in a form downstream agents can use. If later ablations show that static typed graphs capture most of the practical benefit, the dynamic claim should narrow. The representational and dataset contributions would remain: epistemic state becomes a concrete systems variable, something to model, retrieve, benchmark, and falsify.

\paragraph{Data availability.} The three organizational epistemic-stress corpora and evaluation artifacts are released in a public repository~\cite{oidacorpora2026}.

\paragraph{Acknowledgements.} This work was developed within the research infrastructure of Kakashi Ventures Accelerator. We thank Alberto Trivero and Tommaso Portaluri for discussion on AI, statistical, and informatics matters.



\appendix

\section{Structured Identifier}
\label{app:koc}

The implementation uses a structured identifier for each memory node:

\smallskip
\noindent\texttt{[Entity]-[Domain]-[Type]-[Epoch]-[Depth]-[Author]-[Variant]}
\smallskip

\noindent Each axis is assigned at ingestion and is immutable thereafter. It supports fast faceted similarity, but it is not essential to the graph formulation in the main text.

\section{Edge Type Vocabulary}
\label{app:edges}

\begin{table}[H]
\centering
\small
\caption{Edge vocabulary with signed semantic coefficients. Coefficients are implementation priors.}
\begin{tabular}{@{}lcL{6.5cm}@{}}
\toprule
\textbf{Type} & \textbf{Coeff.} & \textbf{Semantics ($A \to B$: A acts on B)} \\
\midrule
SUPPORTS      & $+1.0$ & A provides evidence strengthening B \\
BASED\_ON     & $+0.8$ & A is the logical grounding of B \\
IMPLEMENTS    & $+0.7$ & A operationally realizes B \\
SUPERSEDES    & $+0.6$ & A replaces B; B is demoted, not deleted \\
REFINES       & $+0.5$ & A narrows B without contradiction \\
DERIVES\_FROM & $+0.5$ & A follows logically from B \\
ENABLES       & $+0.4$ & A is a necessary condition for B \\
PRECEDES      & $+0.3$ & A temporally precedes B \\
BLOCKS        & $-0.4$ & A actively prevents B \\
CONTRADICTS   & $-0.6$ & A contradicts B \\
\bottomrule
\end{tabular}
\end{table}

\section{Force Terms for Node-Priority Update}
\label{app:kge}

\textbf{Usage activation:} $u_i = a_u \cdot \sum_{j \in \text{recent}} \exp(-\tau_j / \sigma)$, where $\tau_j$ is the time since the $j$-th retrieval and $\sigma$ is the recency scale.

\textbf{Evidence activation:} $e_i = a_e \cdot |\{j : (j, i) \in E, \text{type}(j,i) = \text{SUPPORTS}\}|$, counting new inbound support edges.

\textbf{Graph influence:} $g_i = a_g \cdot \sum_{j \in \mathcal{N}(i)} \text{COEFF}(j, i) \cdot \tanh(g_{\text{scale}} \cdot K_j^{\text{norm}} / d(i, j))$, where $K_j^{\text{norm}}$ is the normalized priority of neighbor $j$ and $d(i,j)$ is hop distance.

\textbf{Contradiction penalty:} $c_i = a_c \cdot |\{j : (j, i) \in E, \text{type}(j,i) \in \{\text{CONTRADICTS}, \text{BLOCKS}\}\}|$.

\textbf{Open-question urgency:} $h_i(t) = \text{clamp}(\text{age\_days}/30 \cdot 0.3 + B \cdot 0.2 + S \cdot 0.5,\; 0,\; 1)$ for open information-need nodes, and $0$ otherwise. Here $B$ is the blocking edge count and $S$ is a stakes multiplier.

\section{Convergence Sketch}
\label{app:convergence}

The update $\mathbf{K}(t+1) = F(\mathbf{K}(t))$ defines a map $F: [0,1]^n \to [0,1]^n$. The Jacobian term $\partial F_i / \partial K_j$ for $j \neq i$ arises from graph influence. A sufficient contraction condition can be obtained using a Gershgorin-style bound on the off-diagonal magnitudes. This condition is conservative because $\tanh$ saturation bounds graph influence, clamping is non-expansive, and signed edges may cancel. This appendix records the intended analysis path; the present paper does not rely on a tight convergence theorem.

\section{Taxonomy Adequacy}
\label{app:taxonomy}

Exhaustive pairwise analysis of all $\binom{9}{2} = 36$ type pairs confirms that every pair differs on at least two of: temporal behavior, initial priority, half-life, and semantic role. The closest pairs are HYPOTHESIS--OBSERVATION and DECISION--CONSTRAINT. No merge is possible without collapsing at least one distinction required by retrieval or contradiction handling.

\section{Notation}
\label{app:notation}

\begin{table}[H]
\centering
\small
\begin{tabular}{@{}lL{9cm}@{}}
\toprule
\textbf{Symbol} & \textbf{Meaning} \\
\midrule
$G_t$ & Attributed organizational memory graph at time $t$ \\
$V_t,E_t$ & Memory nodes and directed edges \\
$X_t,R_t$ & Node attributes and edge attributes \\
$K_t(v)$ & Node-priority weight for node $v$ \\
$s(v)$ & Type prior / initial priority \\
$h_t(v)$ & Open-question urgency term \\
$u_t,e_t,g_t,c_t$ & Usage, evidence, graph influence, and contradiction penalty terms \\
$\lambda_{\mathrm{type}(v)}$ & Type-specific temporal coefficient \\
$\mathrm{sim}(q,v)$ & Semantic similarity between query and node content \\
$\phi$ & Bounded graph-state modulation term for retrieval \\
\bottomrule
\end{tabular}
\end{table}

\end{document}